\documentclass[conference,a4paper]{APSIPA2024}
\usepackage{amsmath}
\usepackage{graphicx}
\usepackage{multirow}
\usepackage{threeparttable}
\usepackage[backend=biber,style=ieee,]{biblatex}

\usepackage{geometry}
\usepackage{fancyhdr}

\fancypagestyle{firststyle}{
  \fancyhf{}
  \fancyhead[C]{2024 Asia Pacific Signal and Information Processing Association Annual Summit and Conference (APSIPA ASC)}
}

\begin{document}

\title{A Novel LLM-based Two-stage Summarization 
Approach for Long Dialogues}

\author{
\authorblockN{
Yuan-Jhe Yin\authorrefmark{1}, 
Bo-Yu Chen\authorrefmark{2}, and
Berlin Chen\authorrefmark{1}
}

\authorblockA{
\authorrefmark{1}
National Taiwan Normal University, Taiwan}

\authorblockA{
\authorrefmark{2}
National Taiwan University of Science and Technology, Taiwan}

\authorblockA{
E-mail: Yjyjames@gmail.com, patpatpat1015@gmail.com, berlin@ntnu.edu.tw}
}

\maketitle
\thispagestyle{firststyle}
\pagestyle{fancy}

\begin{abstract}
Long document summarization poses a significant challenge in natural language processing due to input lengths that exceed the capacity of most state-of-the-art pre-trained language models.
This study proposes a hierarchical framework that segments and condenses information from long documents, subsequently fine-tuning the processed text with an abstractive summarization model. Unsupervised topic segmentation methods identify semantically appropriate breakpoints.
The condensation stage utilizes an unsupervised generation model to generate condensed data, and our current experiments employ ChatGPT (v3.5).
The summarization stage fine-tunes the abstractive summarization model on the condensed data to generate the final results.
This framework enables long documents to be processed on models even when the document length exceeds the model's maximum input size.  
The exclusion of the entire document from the summarization model reduces the time and computational resources required for training, making the framework suitable for contexts with constrained local computational resources.
\end{abstract}

\section{Introduction}
Current summarization tasks are predominantly classified into extractive and abstractive summarization.
Extractive summarization treats the task as a classification problem, determining which sentences in the article should be included in the summary.
This method ensures that the summary accurately reflects the original article's meaning, preventing misinterpretations and potential hallucinations (factually incorrect or nonsensical content generated by language models).
However, the direct extraction of summary content from the original text may hinder readability and present challenges in attaining high similarity scores compared to human-written labels during performance evaluation.

Abstractive summarization, on the other hand, employs language generation models to create summary content based on the input text.  
This approach often yields summaries that are more coherent and closely align with human-written labels. Nevertheless, the main challenges include achieving model convergence and preventing the generation of hallucinations or inaccuracies. 

For longer text inputs, many studies utilize graph-based models to handle the summarization task \cite{[1],[2],[3],[4]}.  
By constructing graphs to represent the relationships between sentences, these models allow sentences that are relatively distant from each other in the text to influence one another.  
This approach mitigates the adverse effects of distance on summary content. During training, these models necessitate substantial VRAM, and as the input document length increases, the demand for computational resources becomes even more stringent.

A further challenge in summarizing lengthy documents occurs when the article length surpasses the model's input capacity, rendering effective model training impossible.
Certain studies \cite{[5]} tackle this issue through text segmentation techniques that partition articles into smaller, manageable segments.  
\cite{[6]} employs a sliding window to encapsulate information throughout the document.  
Unlimiformer \cite{[7]} modifies the Transformer architecture to accommodate longer input sequences.

We propose an architecture for preprocessing long articles, facilitating their training within commonly used language models.  
Prior research GoSum \cite{[8]} indicates that segmentation is crucial for effective summarization.  
Extracting structural information from text enhances summary quality. Thus, our approach segments lengthy articles to prevent semantic discontinuity during text division, thereby ensuring quality of the summary. 
We utilizes the unsupervised segmentation method outlined in \cite{[9]} for this architecture.
This approach employs the TextTiling algorithm \cite{[10]} and utilizes BERT embeddings to compute similarity, resulting in segmentation outcomes.
Next, we employ zero-shot prompting with a large language model to condense the segmented text, allowing applicability to any dataset irrespective of golden labels. Subsequently, we fine-tune a Transformer-based abstractive summarization model to generate the final summary. The preceding segmentation and condensation steps facilitate processing and training with established summarization models.

The advantage of this architecture is that it enables efficient summarization of lengthy articles with reduced computational resources.
When paired with prompts, this architecture yields superior outcomes on the ForeverDreaming dataset \cite{[11]} . Furthermore, tailored prompts can be utilized based on the dataset and domain to enhance summary quality during the condensation stage.

\section{Related Works}
\subsection{Unsupervised Topic Segmentation}
The method outlined in \cite{[10]} is widely employed for topic segmentation, identifying segment boundaries through shifts in word distribution. The process involves tokenization, determining lexical scores, and identifying boundaries to delineate topical segments.
Building on this method, \cite{[9]} utilized BERT embeddings for sentence representation, achieving improved topic segmentation scores through incorporating BERT's strong semantic representation. Other studies focus on developing superior scoring models for unsupervised topic segmentation. For instance, \cite{[12]} aims to enhance coherence scoring by training on diverse utterance pairs and defining relative scores. Additionally, \cite{[13]} computes relevance scores by integrating topic similarity and coherence scores generated by trained topic and coherence encoders.

\subsection{Dialogue Summarization}

Dialogue summarization is an essential task that includes data types such as meetings, lectures, conversations, and scripts.
The goal is to extract semantic information from dialogue content and succinctly present the key points.

DialogLM \cite{[14]} proposed a pre-training framework featuring a window-based denoising approach that incorporates speaker masking, turn splitting and merging, text infilling, and turn permutation of the original dialogue. 
This framework enhances the model's comprehension of dialogue.
Furthermore, they implemented Sinkhorn attention \cite{[15]} in specific Transformer layers, enhancing local and global information interaction, thereby improving the model's suitability for long dialogue data.
The framework is applied to UniLMv2 \cite{[16]} and Longformer-Encoder-Decoder \cite{[6]} in the study. 

\cite{[17]} proposes self-supervised methods that are similar yet distinct from DialogLM \cite{[14]} for training a summarization model. 
For an input dialogue, interlocutors and utterances are alternated, with irrelevant utterances from other dialogues incorporated into the original dialogue.
In the context of masking interlocutors, only the reference summary is utilized.
A pre-trained BERT \cite{[18]} is fine-tuned to detect modifications in the original dialogue.  
The weights are then shared to initialize a traditional encoder-decoder model, which is subsequently fine-tuned on an abstractive summarization task. 

Some studies concentrate on augmenting the summarization task with supplementary information. 
In dialogues, the speaker's intentions are often implicit, posing challenges for models to fully comprehend the content.
Thus, \cite{[19]} suggests integrating commonsense knowledge to enhance the model's dialogue summarization.
The presented framework utilizes commonsense knowledge model COMET \cite{[20]} and PARA-COMET \cite{[21]} to augment the input dialogue with additional knowledge.  
This augmented input is then fine-tuned using a pre-trained BART \cite{[22]} model. 

\cite{[23]} proposes fusing static graphs representing dialogue discourse relations, keyword co-occurrences, speaker relations, and utterance positions.
The static graphs are integrated with a dynamic graph created via a multi-head attention mechanism. 
This fused graph is used to generate summaries.  
This approach integrates both static information and the dynamic information potentially absent in static graphs.

\subsection{Long Document Summarization}

In long document summarization, two primary challenges are the document's input length exceeding the model's limit and the difficulty of capturing its structured information.

\cite{[24]} addresses this issue by altering the cross-attention computation in Transformer architectures to enable unlimited input. Input data are encoded in chunks and stored in an index in GPU or CPU memory. In each cross-attention computation, the k-nearest neighbors of the query are selected from the index for inclusion in the computation. This approach guarantees the availability of each input sequence chunk for cross-attention computation.

In contrast to \cite{[25]}, which constructs a heterogeneous graph using word and sentence nodes, GoSum \cite{[8]} builds the graph with sentence and section nodes, utilizing a reinforcement learning-based extractive summarization model. In this structure, sentence nodes connect to their respective section nodes, with both sentences and sections fully interconnected. This configuration conveys hierarchical semantic information and mitigates semantic drift across sections. Eliminating word nodes reduces the model's runtime and enhances the structural representation of lengthy documents.

\cite{[24]} introduced a hierarchical propagation layer to facilitate information dissemination across multiple Transformer windows. The input data is segmented into multiple blocks and tokenized. The propagation layer initiates with a Transformer, subsequently processing the CLS token of each block through a BIGRU network to aggregate semantic information across blocks. This architecture enables BERT to handle input data exceeding its maximum size while retaining information from the entire document via updates to the CLS representation.

The study \cite{[26]} introduces three methods for long document summarization: 'Direct', 'Chunk and Summarize', and 'Extract then Summarize'. The Direct approach omits input data preprocessing, with automatic truncation to the model's maximum token limit. The 'Chunk and Summarize' approach segments the input document at the maximum token limit, processes each segment independently, and merges the resulting summaries. Another variant entails inputting the generated summary with the subsequent segment into the model. The Extract then Summarize approach generates sentence embeddings for dialogues and summaries, subsequently calculating cosine similarity. A similarity threshold is established to exclude sentences with low similarity. Ultimately, the extracted dialogues are input into the models to generate summaries. The experiment utilized several state-of-the-art summarization models, including Longformer \cite{[6]}, T5 \cite{[27]}, Flan T5 \cite{[28]}, BART \cite{[22]}, and Chat-GPT, and was evaluated on datasets such as Qmsum \cite{[25]} and ForeverDreaming \cite{[11]}. Since we are utilizing ForeverDreaming \cite{[11]} in our evaluation experiments, its results provide baseline comparisons.

\begin{figure*}[!t]
    \centering
    \centerline{\includegraphics[width=1\linewidth]{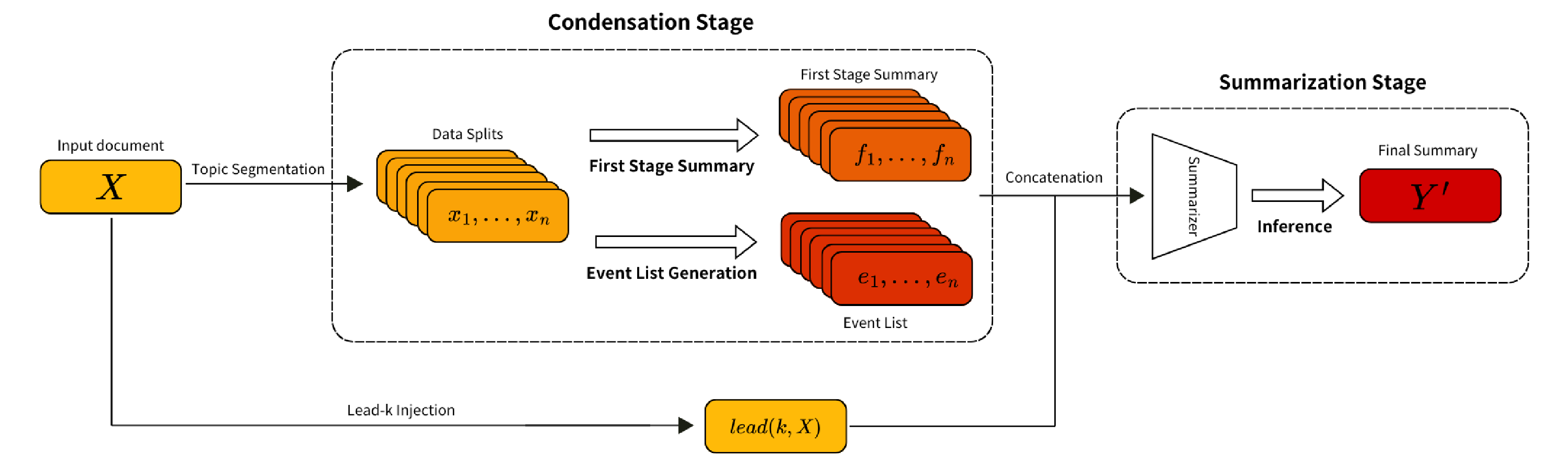}}
    \caption{Workflow of the proposed framework.}
    \label{fig:res}
\end{figure*}

\section{Proposed Framework}
We propose a hierarchical framework for document summarization that involves multiple stages. First, the input document is divided into segments based on semantic information. These segments are then condensed using zero-shot prompting to generate a first-stage summary. Next, each first-stage summary is concatenated with $k$ utterances extracted from the original input. Finally, the processed data is fine-tuned using an abstractive summarization model to produce the final summary. 

\subsection{Topic Segmentation}
A single input document consists of $n$ utterances$\ X=\{u_1,u_2,...,u_n\}$. We aim to determine the semantic breakpoints in the document for data splitting. Let semantic breakpoint $sb_j$ be the index of last utterance in the $j$th segment, then we will have segment index set	$SEG=\{sb_1,sb_2,...,sb_l\}$, with $l$ segments .
We referenced the approach proposed in \cite{[9]} and adjusted it according to our requirements and experimental results. Unlike \cite{[9]}, our approach does not apply the block-wise max pooling operation. Instead, a sliding window method is used for BERT embedding calculation. Specifically, the embedding of $i$th utterances $eb_i$ is calculated by concatenating the previous and next utterances:$$eb_i=tokenizer(u_{i-1}\oplus u_i\oplus u_{i+1})$$ $$sim_i=cos(eb_i\cdot eb_{i+1})$$ where $cos(\cdot)$ denotes cosine similarity  function and $sim_i$ denotes the cosine similarity between the BERT embeddings of the utterances  $u_i$ and $u_{i+1}$. 

\cite{[9]}'s work specifies the semantic breakpoint to be determined by every $sim_i$ such that: $$sim_i<\mu_S-\sigma_S$$ where $\mu_S , \sigma_S$ is the mean and variance of all cosine similarities  $S=\{sim_1,sin_2,...sim_n\}$. This method resulted in too many semantic breakpoints when the variance was too small, leading to excessive segmentation and loss of meaningful context. Therefore, we adopted a greedy approach for selecting semantic breakpoints that is more-suitable for our experiment. The steps are as follows: 1. Separate the utterance with the least cosine similarity as a semantic breakpoint. 2. Protect the nearby $w$ utterances from segmentation. 3. Repeat steps 1 and 2 until the specified number of segments $l$ is reached or no suitable breakpoint can be found.

$w$ and $l$ are customizable parameters. By protecting adjacent sentences from being split and allowing for an adjustable number of segments, the algorithm can better adapt to various scenarios.

\subsection{Data Splitting}
To accommodate the maximum input size of the summarization model, we perform a data split based on the results of topic segmentation, ensuring that semantic meaning is preserved. The pseudo segment size 
$M$ is determined by the model's maximum input capacity. Starting from every $M$ utterances in the input text, we locate the nearest semantic breakpoint. This process is conducted backwards to ensure that the length of each segment does not exceed $M$ utterances.
The $i$th data split denotes $x_i=\{u_{sb_j+1},u_{sb_j+2},...,u_{sb_{k}} \}$, where $j$ and $k$  are some index in segment index set such that the length of utterances in segment $j+1$ to $k$ is not greater than $M$. 
After data splitting, the input document $X=\{x_1\cap x_2\cap ... \cap x_m\}$ consists of $m$ splits.

\subsection{Condensation Stage}
Performing summarization tasks using zero-shot prompting with ChatGPT may result in summaries presented from different perspectives or overly vague summaries (see the analysis section). Consequently, the generated summaries can deviate from the golden labels. By combining the first-stage summary with the event list method, our model can separately generate text with a summarization structure and neutral perspective descriptions. This approach allows the model in the summarization stage to obtain more comprehensive information.
We use ChatGPT (v3.5) to generate the first-stage summary of the input document, ensuring that its length meets the input length requirements of the BART model. The first-stage summary generation process is as follows:
$$f_i=LLM(x_i,p_s)$$$$F=f_1\oplus f_2 \oplus ... \oplus f_m$$where $f_i$ is the first-stage summary of $x_i$ , $p_s$ denotes the prompt used to generate first-stage summary, and $LLM()$ denotes the zero-shot prompting model. The generated results are concatenated to form the first-stage summary $F$ corresponding to input document $X$.

The event list generation process is as follows:
$$e_i=LLM(x_i,p_e)$$$$E=e_1\oplus e_2 \oplus ... \oplus e_m$$where $e_i$ is the event list of $x_i$ , $p_e$ denotes the prompt used to generate event list, and $E$ is the generated event list for $X$.

By generating the first-stage summary and event list, the information from the input document is condensed to a trainable length suitable for summarization models like BART.

\subsection{Summarization Stage}
To further enrich the training data, we inject the first $k$ utterances of the original text into the training data. This ensures that the content processed by the LLM remains closely aligned with the original material, thereby enhancing summarization performance. We define enhanced input data as: $$X'=E\oplus F \oplus lead(k,X) $$where $lead(k,X)$ function extracts first $k$ utterances from $X$.
Finally, we use the enhanced input data from the training set to train a BART model. Let the golden summary be $Y=\{y_1,y_2,...,y_k\}$ , where $y_i$ is the $i$th utterance of  $Y$. The model minimizes the cross-entropy loss as follows: 
$$L=-\sum_{t=1}^T log P(y_t|y_{<t},X')$$ where  $y_t$ is the actual token at position $t$, $y_{<t}$   represents all tokens before position $t$, $X'$ denotes the enhanced input data. Note that the enhanced input is also applied during the inference phase; the test set data will be processed in the same manner to generate the result summary.

\section{Experiment}
\subsection{Dataset}
We conducted experiments on the ForeverDreaming dataset, which includes TV show transcripts paired with human-written summaries. The dataset contains 4,348 dialogues, comprising 3,673 training samples, 338 validation samples, and 337 test samples.

\subsection{Baselines}
We referenced the experimental data from the study \cite{[5]}, ensuring that the experimental details are consistent. The experiments were conducted on a randomly selected 15\% subset of the ForeverDreaming dataset. Several approaches were presented, including 'Direct', 'Chunk and Summarize', and 'Extract then Summarize'. For detailed methods, please refer to the related works section. We present the result with the best ROUGE-1 score under their experimental settings for comparison. For the comprehensiveness of the baseline data, we also referenced DialogLM's results on the ForeverDreaming dataset. It is important to note that both our experimental results and those from DialogLM were obtained using the complete ForeverDreaming dataset.

\textbf{BART}: BART is a state-of-the-art denoising sequence-to-sequence pre-trained model, often used in natural language generation and translation. The BART-2048 model with the 'Chunk and Summarize' approach is used.

\textbf{Longformer}: Longformer replaces the standard self-attention mechanism for long document input, making it suitable for generative sequence-to-sequence tasks involving long documents. The Longformer model with the 'Extract then Summarize' approach is used.

\textbf{T5}: The Text-to-Text Transfer Transformer (T5) converts every language problem into a text-to-text format, commonly used for summarization tasks. The T5-base-2048 model with the 'Chunk and Summarize' approach is used.

\textbf{ChatGPT}: ChatGPT is a conversational generation model developed by OpenAI, based on the GPT-3 and GPT-4 architectures. It is designed for natural language interactions and provides high-quality exchanges in various application scenarios. ChatGPT (V3.5) with the 'Direct' approach is used.

\textbf{DialogLM}: DialogLM is a state-of-the-art pre-trained summarization model that focuses on long dialogue. Its window-based denoising method enables the model to gain a better understanding of long documents.

\subsection{Experimental Results}
\begin{table}
    \centering
    \caption{Experiment Results and Baseline}
    \label{tab:t1}
    \begin{tabular}{cccc}
        \hline
        Summarization & ROUGE-1 & ROUGE-2 & ROUGE-L\\
         \hline
         Our model  & \textbf{29.15} &  \textbf{6.72 }&  \textbf{24.54}\\
         Baseline &  &  & \\
         BART &  19.35 & 2.02  & 17.28  \\
        LongFormer&  15.39 & 1.42  &12.52  \\
         T5  & 18.57 & 1.48  &17.62   \\
         ChatGPT &  25.04 & 4.34 &18.31 \\
         DialogLM-large &  \textbf{36.70} & \textbf{8.38} &\textbf{31.38} \\
          \hline
    \end{tabular}
\end{table}
Our framework outperforms the results obtained by \cite{[5]}. The 'Chunk and Summarize' method is similar to ours in that it adopts the idea of segmenting input data. However, our approach extends further. By using an unsupervised topic segmentation method, the data splitting aligns more closely with the textual information. We then use the event list generation method to further integrate the segmented data, and finally, we fine-tune the summarization results using BART. This approach allows us to achieve a difference of more than 9 points in ROUGE-1 score compared to 'BART'.

The 'ChatGPT' result in the table does not use segmentation and performs better compared to 'Summary Generation' in Table 3. This indicates that using segmentation with ChatGPT may result in a lower summary score. ChatGPT achieves the best results when the article is not segmented. However, by employing event list generation and fine-tuning, we can achieve scores that surpass those of ChatGPT summarization.

DialogLM achieves better ROUGE scores in the ForeverDreaming summarization task. The primary reason is the approach utilizes the full text of long documents and involves a relatively long pre-training process. The complete information is fed to the model instead of condensed text, requiring significant computational resources and training times. In contrast, our method requires fewer computational resources to achieve comparable results, making long document summarization feasible for computers with less VRAM.

\section{Analysis}
\subsection{ChatGPT Summarization}
\label{ssec:cs}
\begin{table}
    \centering
    \caption{Comparison of ChatGPT Summarization}
    \label{tab:t2}
    \begin{tabular}{cccc}
        \hline
        Methods & ROUGE-1 & ROUGE-2 & ROUGE-L\\
         \hline
         Event List Generation &  22.47 & 3.39  & 19.69 \\
         First Stage Summarization & 21.99 & 3.11 & 19.43   \\
         ChatGPT Second Stage Summarization  & 20.52 & 2.87   &17.67     \\
         BART Second Stage Summarization &  27.27 & 5.33 &22.90 \\
          \hline
    \end{tabular}
\end{table}

To assess the impact of different prompts on summary content, we conducted further analysis on results generated by the LLM during the zero-shot prompting data condensation stage. The result is presented in Table II.
The 'Event List Generation' and 'First-Stage Summary' rows indicate the ROUGE scores of the concatenated event list $E$ and the first-stage summary $F$ compared to the golden label.
The 'Second Stage Summarization' rows represent the further summarization of the concatenated first-stage summary.

The results for event list summarization marginally surpass those for direct text summarization.
This discrepancy arises from the 'summarizing' prompt potentially producing vague or subjective summaries.
Modifying the prompt to "event list" yields a text comprised of objective descriptions of the events, thereby more accurately reflecting the article's content.

The effects of ChatGPT summarization are evident in the 'First Stage Summarization' and 'ChatGPT Second Stage Summarization' rows.
The same prompts were used to generate both results. 
The ROUGE score decreased with multi-stage zero-shot prompting summarization, indicating that while this method can be used to condense and shorten the original document, the information loss issue remains unsloved.

The results from the 'BART Second Stage Summarization' indicate that fine-tuning substantially enhances the ROUGE score, justifying our selection of BART as the final summarization model. 
Moreover, integrating 'Event List Generation' and 'First Stage Summarization' as inputs enables the model to capture the summarization structure and neutral descriptions of the original documents, thereby enhancing outcomes.

\subsection{Lead-$k$ Injection}
\label{ssec:lk}

\begin{table}
    \centering
    \caption{Compirsion of Lead-$k$ Injection}
    \label{tab:t3}
    \begin{tabular}{cccc}
        \hline
        Methods & ROUGE-1 & ROUGE-2 & ROUGE-L\\
         \hline
         Lead-0  &  28.96 & 6.58   & 24.40\\
         Lead-1  & 29.02  & 6.59   & 24.45   \\
         Lead-3   & 28.38  & 6.32   & 23.80     \\
         Lead-5  &  \textbf{29.15} &  \textbf{6.73}&  24.54 \\
         Lead-10   & 28.97  & 6.72   & \textbf{24.55}     \\
          \hline
    \end{tabular}
\end{table}

The Lead-$k$ Injection method also helps the model gain a better understanding of the original document, mitigating information loss during data preprocessing. To further understand the effects of Lead-$k$ Injection, we tested how different values of $k$ for injection affect the ROUGE score. Lead-0 indicates no injections, meaning the model is trained on the event list only. The table shows that Lead-$k$ Injection enriches the input text, resulting in a better ROUGE score. However, excessively increasing $k$ may lead to a loss of focus during the training phase, thereby decreasing model performance.

\subsection{Summarization Model}
\label{sec:sp}

\begin{table}
    \centering
    \caption{Compirsion for different Summarization Model}
    \label{tab:t4}
    \begin{tabular}{cccc}
        \hline
        Model & ROUGE-1 & ROUGE-2 & ROUGE-L\\
         \hline
         Pegasus  &  25.57 & 5.20  & 21.59 \\
        BART & 28.96 & 6.58   & 24.40  \\
          \hline
    \end{tabular}
\end{table}

In the Sammarization Stage, different summarization models can be applied to the framework. Experiments were conducted using both Pegasus and BART for comparison. Note that the presented results are generated with Lead-0 (no utterances injected), as the optimal $k$ may vary for each model. The results show that BART outperforms Pegasus by approximately 6 points in ROUGE-1 score, indicating that BART is the superior model for our task.

\section{Conclusion}

\label{sec:cc}

The proposed framework effectively summarizes long documents with low computational cost. 
Despite the data condensation process shortening the original document, some essential information may be lost.
Currently, we use a naive solution that injects the first $k$ utterances into the abstractive summarization model input. 
This method can be enhanced by employing ROUGE-based selection for the injected segment or by fine-tuning an extractive summarization model to identify and preserve key information from the original document.
Summarizing long documents, such as those in the ForeverDreaming dataset, using common Transformer-based models, demands significant VRAM and extensive training time.
Despite the additional costs of using the ChatGPT API, the framework is viable for scenarios lacking local computational resources.

\end{document}